\documentclass[letterpaper]{article} 
\usepackage{aaai23}  
\usepackage{times}  
\usepackage{helvet}  
\usepackage{courier}  
\usepackage[hyphens]{url}  
\usepackage{graphicx} 
\usepackage{natbib}  
\usepackage{caption} 
\usepackage{algorithm}
\usepackage{algorithmic}
\usepackage{algorithm}
\usepackage{algorithmic}
\usepackage{amsmath}
\usepackage{color}
\usepackage{booktabs}

\usepackage{setspace}
\usepackage{multirow}
\usepackage{makecell}

\usepackage{amssymb}
\usepackage{amsmath}
\usepackage{bm}
\usepackage{newfloat}
\usepackage{listings}
\DeclareCaptionStyle{ruled}{labelfont=normalfont,labelsep=colon,strut=off} 
\floatstyle{ruled}
\newfloat{listing}{tb}{lst}{}
\floatname{listing}{Listing}
\title{IKOL: Inverse kinematics optimization layer for 3D human pose and shape estimation via Gauss-Newton differentiation}
\author {
    Juze Zhang \textsuperscript{\rm 1,\rm 2,\rm 3,\rm 4},
    Ye Shi \textsuperscript{\rm 1,\rm 4}\equalcontrib,
    Yuexin Ma \textsuperscript{\rm 1,\rm 4},
    Lan Xu \textsuperscript{\rm 1,\rm 4},
    Jingyi Yu \textsuperscript{\rm 1,\rm 4},
    Jingya Wang \textsuperscript{\rm 1,\rm 4}\equalcontrib
}
\affiliations {
    \textsuperscript{\rm 1} ShanghaiTech University 
    \textsuperscript{\rm 2} Shanghai Advanced Research Institute, Chinese Academy of Sciences\\
    \textsuperscript{\rm 3} University of Chinese Academy of Sciences
    \textsuperscript{\rm 4}  Shanghai Engineering Research Center of Intelligent Vision and Imaging \\
    \{zhangjz,shiye,mayuexin,xulan1,yujingyi,wangjingya\}@shanghaitech.edu.cn
}

\usepackage{bibentry}

\begin{document}

\maketitle

\begin{abstract}
This paper presents an inverse kinematic optimization layer (IKOL) for 3D human pose and shape estimation that leverages the strength of both optimization- and regression-based methods within an end-to-end framework. IKOL involves a nonconvex optimization that establishes an implicit mapping from an image’s 3D keypoints and body shapes to the relative body-part rotations. The 3D keypoints and the body shapes are the inputs and the relative body-part rotations are the solutions. However, this procedure is implicit and hard to make differentiable. So, to overcome this issue, we designed a Gauss-Newton differentiation (GN-Diff) procedure to differentiate IKOL. GN-Diff iteratively linearizes the nonconvex objective function to obtain Gauss-Newton directions with closed form solutions. Then, an automatic differentiation procedure is directly applied to generate a Jacobian matrix for end-to-end training. Notably, the GN-Diff procedure works fast because it does not rely on a time-consuming implicit differentiation procedure. The twist rotation and shape parameters are learned from the neural networks and, as a result, IKOL has a much lower computational overhead than most existing optimization-based methods. Additionally, compared to existing regression-based methods, IKOL provides a more accurate mesh-image correspondence. This is because it iteratively reduces the distance between the keypoints and also enhances the reliability of the pose structures. Extensive experiments demonstrate the superiority of our proposed framework over a wide range of 3D human pose and shape estimation methods. Code is available at https://github.com/Juzezhang/IKOL.
\end{abstract}

\begin{figure}[t]
\centering
	\includegraphics[width=1.0\columnwidth]{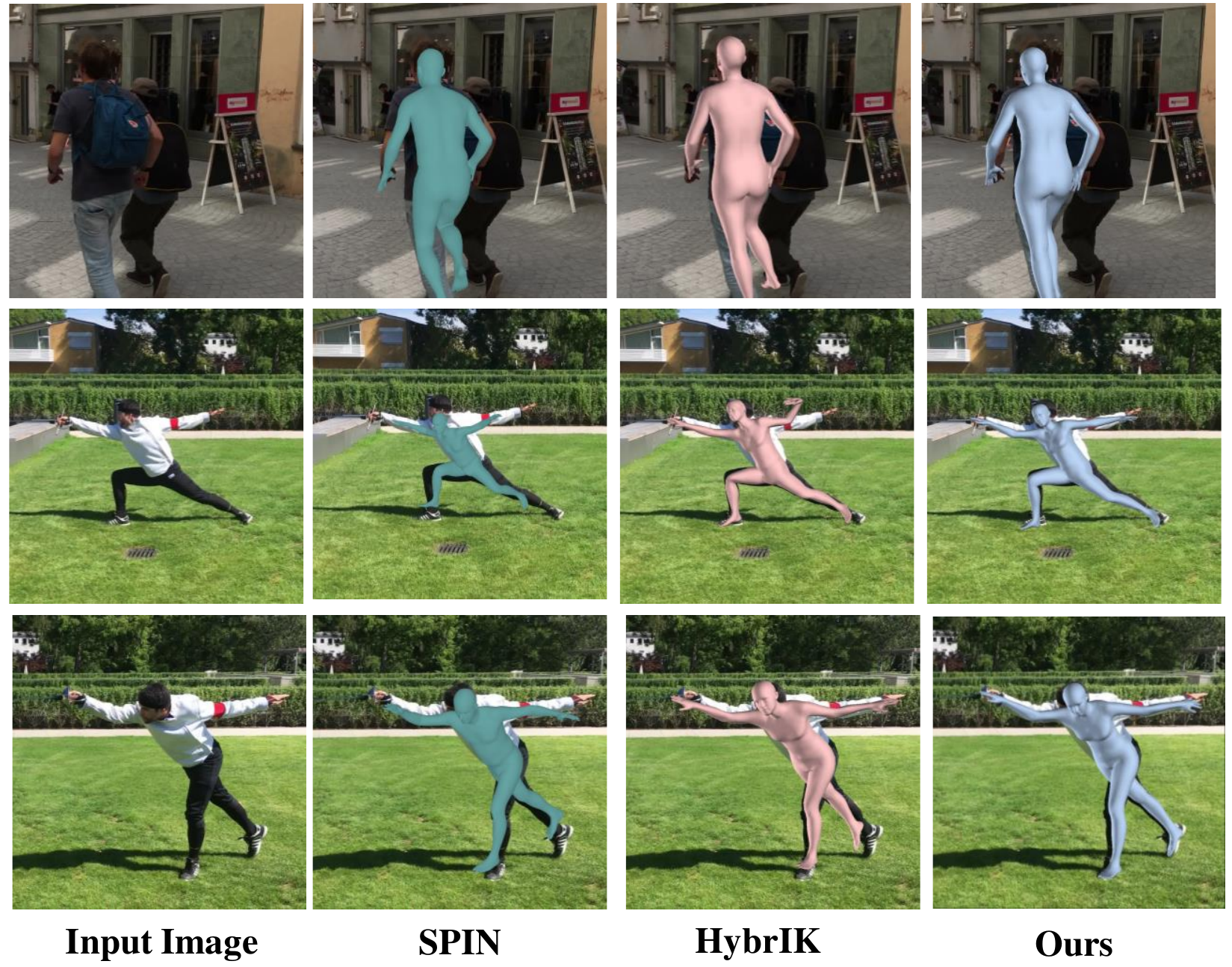}
\caption{Qualitative comparison with the state-of-the-art method SPIN \cite{kolotouros2019learning} and HybrIK \cite{li2021hybrik}.}
\label{Overview}
\end{figure}

\section{Introduction}
Recent years have seen the widespread application of 3D human pose and shape (3D-HPS) estimation to a range of fields, including video analysis, camera surveillance, human computer interaction, virtual/augmented reality, and others. Among the leading techniques, two popular paradigms for 3D-HPS have been investigated: optimization-based methods and regression-based methods. Optimization-based methods \cite{bogo2016keep, fan2021revitalizing} iteratively fit parametric models (such as SMPL \cite{loper2015smpl}) to 2D observations. They can produce accurate mesh-image alignments, and they can also preserve consistency between some input data and a prediction. However, fitting the parametric models to 2D observations is an inherently ill-posed problem; it is also non-convex. This means various optimization techniques are required to find a local minimum and compute an initial guess. Worse still, this procedure is time-consuming and the final accuracy of the estimation relies heavily on the initial guess. Conversely, regression-based methods treat 3D-HPS as a regression problem, where a deep network is trained to directly regress the model’s parameters thanks to the powerful nonlinear mapping capability of neural networks. However, despite efficient and promising results, regression-based methods tend to suffer from misalignments between the estimated meshes and the image. 

Some pioneering studies have attempted to integrate these two paradigms, coupling an optimization process with a regression technique  \cite{zanfir2021neural,guler2019holopose,kolotouros2019learning}. These works either use post-optimization schemes to refine the regression results during the inference stage, or they use optimization to provide a supervision signal to the optimization scheme in the training loop. The former approaches (\cite{zanfir2021neural,guler2019holopose}) tends to rely heavily on the quality of the regression result, while the latter approach (SPIN, \cite{kolotouros2019learning}) can suffer from an instability during training process due to inaccurate supervision and initialization sensitivity. Thus, to date, trying to leverage the best of both works still requires further investigation. Recently, \cite{li2021hybrik} proposed HybrIK by deriving analytical solutions to handle inverse kinematics via a twist-and-swing decomposition based on the assumption that the bone lengths are consistent. Although this simplification is effective, using this strategy can sacrifice some accuracy of the inverse kinematics. This is because assuming that bone lengths are consistent will usually result in non-trivial errors, stemming from potential regression errors and the measurements of the 2D and 3D keypoints. 

To alleviate these issues, we designed an inverse kinematics optimization layer (IKOL) that leverages the strengths of both optimization and regression for end-to-end 3D-HPS estimation. IKOL employs a twist-and-swing decomposition and an optimization process that does not rely on the simplified assumption of bone lengths. In contrast to SPIN, IKOL’s optimization process of is very fast and very effective because it benefits from decomposition. It is worth noting that IKOL involves a nonconvex optimization with the 3D keypoints and body shape as inputs and the relative body-part rotations as the solutions, which is hard to make differentiable. In recent years, some solvers such as Optnet \cite{amos2017optnet}, CvxpyLayer \cite{agrawal2019differentiable} and Alt-Diff \cite{sun2022alternating} have been developed to differentiate convex optimizations. Although these solvers have achieved considerable success, differentiation schemes for nonconvex optimizations are still quite open. The problem with 3D-HPS estimation is that it is highly nonlinear and also nonconvex. This motivated us to develop an efficient differentiation procedure. Called Gauss-Newton differentiation (GN-Diff), this scheme iteratively linearizes the nonconvex objective function to obtain Gauss-Newton directions with closed form solutions. Thus, an automatic differentiation procedure can be directly applied to generate a Jacobian matrix for end-to-end training. Notably, the GN-Diff procedure is quite fast because it does not rely on a time-consuming implicit differentiation procedure. In summary, our work makes three main contributions to the literature: 
\begin{itemize}
\item IKOL: which leverages the strengths of both the optimization and regression for end-to-end 3D-HPS estimation. IKOL has a much lower computational overhead than most existing optimization-based methods and provides a more accurate mesh-image correspondence than existing regression-based methods. 
\item GN-Diff: which efficiently differentiates the nonconvex optimization in IKOL. Notably, existing solvers of optimization layers cannot differentiate nonconvex optimization problems. In addition, our GN-Diff procedure is very fast because it benefits from iterative closed form solutions via Gauss-Newton iterations. 
\item State-of-the-art performance: IKOL with GN-Diff yields state-of-the-art performance with a wide range of 3D human pose and shape benchmarks. Notably, the time cost of the optimization branch is much faster than the competing optimization procedure used in SPIN and is even competitive with the analytical solution proposed in HybrIk. 
\end{itemize}

\section{Related work}
\subsection{Optimization-based methods}
The current optimization methods usually fit a parametric model (such as SMPL \cite{loper2015smpl}) to image cues. Based on this formulation, several image cues have been designed as fitting terms for object functions, including 2D keypoints \cite{bogo2016keep}, 3D keypoints \cite{mehta2017vnect,mehta2020xnect,zhang2022mutual}, silhouettes \cite{huang2017towards, lassner2017unite}, part segmentation\cite{zanfir2018monocular}, part orientation fields \cite{xiang2019monocular}, and more\cite{han2022licamgait,liang2022hybridcap}. However, only using fitting term may lead to unnatural and unrealistic shapes and poses. Alleviating this issue relies on several prior terms, such as a mixture of Gaussians \cite{bogo2016keep}, a variational autoencoder \cite{pavlakos2019expressive}, or a normalizing flow \cite{zanfir2020weakly}. Yet, despite the good performance of optimization-based methods with image-mesh alignment, most treat the fitting scheme as a separate task. As a consequence, accuracy relies heavily on estimated observations. By contrast, our method formulates the problem as an optimization layer and learns the entire regression and optimization together in an end-to-end manner.

\begin{figure*}[t]
\centering
	\includegraphics[scale = 0.6]{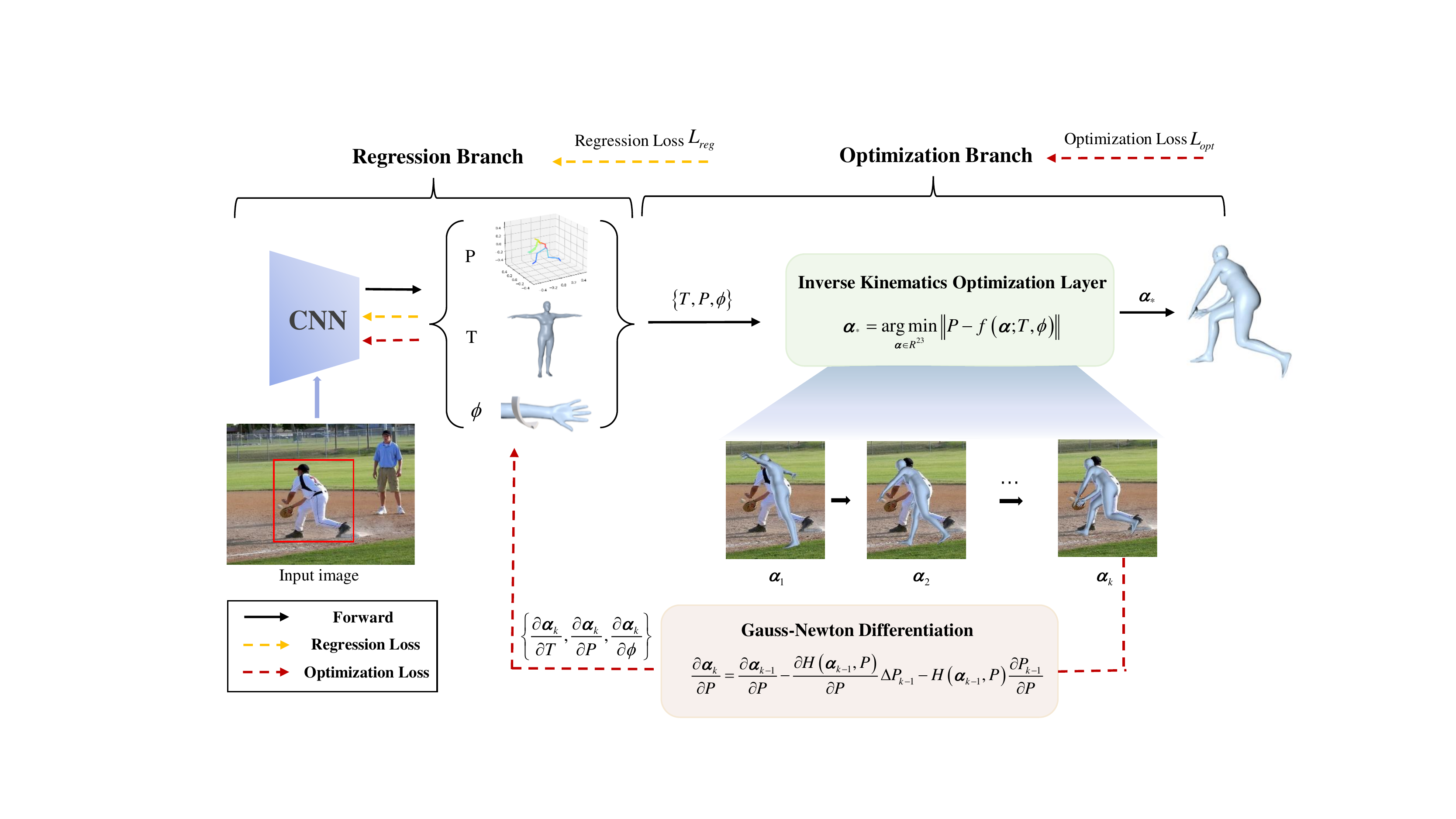}
\caption{\textbf{Overview of the proposed inverse kinematics optimization layer(IKOL) framework.} First, a CNN backbone is utilized to regress the 3D joints $P$, the rest pose $T$ and the twist angle $\phi$. Second, the inverse kinematics optimization layer involves an optimization scheme that establishes the mapping from the outputs of the regression branch to the swing angle $\bm{\alpha}$. IKOL is trained in an end-to-end differentiable framework by using GN-Diff.}
\label{Overview}
\end{figure*}

\subsection{Regression-based methods}
There has been a recent trend to treat 3D-HPS estimation as a regression problem, i.e., where the model’s parameters are directly regressed thanks to the powerful nonlinear mapping capability of deep neural networks. Since the ground truth of the model’s parameters, or the model mesh, is rarely available, some have used 2D annotations as a form of weak supervision by enforcing different kinds of reprojection loss. The 2D annotations used include 2D keypoints\cite{kanazawa2018end,pavlakos2018learning}, silhouettes\cite{pavlakos2018learning, varol2018bodynet}, parts segmentation\cite{omran2018neural, rueegg2020chained} and dense correspondences \cite{xu2019denserac,zhang2020learning}. To overcome occlusion issues, a few studies \cite{zhang2021body,sun2021monocular,sun2022putting, jiang2020coherent} have involved one-stage networks, which simultaneously regress multiple 3D people. Yet, despite their effectiveness, mappings from an image to a parametric model are hard to learn, and so the strategy tends not to preserve accurate image-mesh alignments. By comparison, IKOL involves an optimization that does provide an accurate mesh-image correspondence by iteratively reducing the distance between keypoints. 

\subsection{Hybrid of Optimization \& Regression methods}
Several research teams have tried to integrate optimization processes with regression methods for 3D-HPS estimation. So as to use the iterative fitting strength of an optimization method, some studies have adopted a post-optimization scheme which refines their results after the regression is complete. This ensures the output aligns with the intermediate estimations, such as part segmentation\cite{zanfir2021neural}, and dense correspondences\cite{guler2019holopose}. However, typically, the post-optimization results rely heavily on the quality of the regression results and are sensitive to the initialization. Unlike the above post-optimization methods, SPIN \cite{kolotouros2019learning} performs its optimization in the training loop as a way to provide a supervision signal for the regression branch. However, this type of supervision has a tendency to be inaccurate and is quite sensitive to the initialization settings. Often, this means the training process is unstable. To avoid an iterative fitting process to find the optimal solution, recent studies \cite{li2021hybrik, moon2020i2l, yu2021skeleton2mesh} have suggested building a collaboration between the 3D joints and the body mesh. Although an analytical solution can be produced to replace the optimization procedure by removing the twist components learned from a neural network, this method assumes the bone length consistency hypothesis. Consequently, inverse kinematic accuracy may suffer. In contrast to HybrIK, IKOL formulates the inverse kinematics problem as an iterative optimization process instead of as an approximated analytical derivation. 

\subsection{Optimization Layer}
Embedding the optimization process as a layer in a DNN has emerged as a promising new direction in building models \cite{amos2017optnet, agrawal2019differentiable, gould2021deep,sun2022alternating}. Optnet \cite{amos2017optnet}, for example, integrates quadratic optimization into a DNN and implicitly differentiates the Karush–Kuhn–Tucker (KKT) conditions for an end-to-end training scheme. Subsequently, CvxpyLayer \cite{agrawal2019differentiable} was devised. This framework differentiates a general convex optimization based on conic operators and has been applied to many tasks, such as optimizing rank metrics \cite{rolinek2020optimizing} and graph matching \cite{rolinek2020deep}. However, these methods only work for convex optimization; they are not suitable for the nonconvex problems associated with 3D-HPS estimation. For this reason, we designed GN-Diff, which efficiently does this differentiation work.

\section{Method} 
In this section, we present the IKOL method and the GN-Diff procedure, which together create an end-to-end learning framework for 3D-HPS estimation. Within the model, a regression branch predicts the 3D joints, and the IKOL layer estimates the 3D keypoints to boost the 3D body mesh estimation. Our pipeline for IKOL is illustrated in Fig. \ref{Overview}



\subsection{Inverse Kinematics Optimization Layer}
In this paper, we used SMPL\cite{loper2015smpl} as the body model, which provides a differentiable function $\mathcal{M}(\theta,\beta)$ to control an artist-created mesh with $N=6890$ vertices and $K=23$ joints. The body shape is parameterized by the first 10 principal components $\beta \in \mathbb{R}^{10}$. Following standard skinning practice, the zero pose $\theta^*$ results in the mean template shape $\hat{T} \in \mathbb{R}^{6890}$. The rest pose of joints $T\in \mathbb{R}^{23}$ are obtained via a linear regressor mapping from $\hat{T}$. The relative axis angle of the joints are parameterized by $\theta \in \mathbb{R}^{3 \times 23}$ and usually transformed into a relative rotation matrix $ R = \{R_{par(i),i}\}_{i=1}^K$ using Rodrigues’ formula. Here, $\text{par}(i)$ denotes the parent of the body joint $i$ and $R_{par(i),i}  \in \mathbb{SO}(3)$ denotes the $i$-th local rotation matrix with respect to its parent joint. We denote the $i$th global rotation matrix as $R_i \in \mathbb{SO}(3)$. With this, the global rotation matrix of the body joint $i$ can be recursively computed from relative rotation matrix as:
\begin{equation}
R_i = R_{par(i)}R_{par(i),i}.
\end{equation}
It is worth noting that the rotation estimated from the regressed joints $P$ are inherently ambiguous as the twist angle is missing from the skeleton representation. Hence, we followed \cite{li2021hybrik} and decomposed the rotation $R$ into a twist rotation $R^{tw}$ and a swing rotation $R^{sw}$ by applying twist-and-swing decomposition \cite{baerlocher2001parametrization}. Thus, each rotation matrix can be computed as:
\begin{equation}
R_i = R^{sw}_i(\bm{\alpha}_i) R^{tw}_i(\phi_i)
\label{roation}
\end{equation}
where $\bm{\alpha}_i \in \mathbb{R}$ denotes the swing angle of $i$th rotation that will optimized by our optimization branch and $\phi_i \in \mathbb{R} $ is the twist angle of $i$th rotation that will be predicted by the regression branch. Readers interested in further derivation details of the swing rotation $R^{sw}(\bm{\alpha})$ and twist rotation $R^{tw}(\phi)$ are referred to \cite{li2021hybrik}. Notably, pelvis joints and the leaf joints are excluded in this decomposition. Because the pelvis joints have three children which can be derived by closed-form solution with the children joints of the spine, left hip and right hip. 
Thus, the forward kinematics of SMPL are defined as:
\begin{equation}
Q = FK(R^{sw}(\bm{\alpha}); R^{tw}(\phi), T, P) := f(\bm{\alpha}),
\label{forward}
\end{equation}
where $Q \in \mathbb{R}^{23}$ denotes the reconstructed 3D joints from the SMPL outputs, $P$ denotes the regressed 3D joints estimated from the regression branch, $\bm{\alpha}\in \mathbb{R}^{23}$ denotes the swing angle, and $f(\cdot)$ denotes the forward model operator for the simplification. 

Directly regressing the model parameters is non-trivial due to the highly non-linear mapping between the images cues and the articulated pose. Usually, such a procedure leads to image-model misalignment. By contrast, 3D keypoint representation predicts the per-pixel likelihood for each joint’s location as a 3D heatmap, which can preserve the spatial relationships in the input image. 

Similar to the network design in HybrIK \cite{li2021hybrik}, we also use a deep neural network to predict the 3D joints $P$, the twist angle $\phi$, and the shape parameters $\beta$. We use the common supervision of 3D joints, and shape and twist angles as follows:


\begin{equation}
L_{reg} = w_1 L_{shape} + w_2 L_{twist} + w_3 L_{joint}, 
\end{equation} 
where $L_{shape}$ is the $\ell_2$ loss of the shape parameters $\beta$, $L_{twist}$ is the $\ell_2$ loss of the twist angle $\beta$, $L_{joint}$ is the $\ell_1$ loss of the regressed joints between the groundtruth and $w_{(\cdot)}$ is the corresponding loss weights. Once the regressed joints $P$, the twist angle $\phi$, and the shape parameters $\beta$ have been obtained from regression neural network, IKOL can be used to obtain the corresponding solution angle:

\begin{equation}\label{loss1}
\bm{\alpha}_* = \arg\min \limits_{\bm{\alpha}} \frac{1}{2} \| P-f(\bm{\alpha})\| _2 ^2.
\end{equation} 

Then, the optimization loss is designed to compute the loss with  ground truth of the rotation as follows: 


\begin{equation}
L_{opt} =  \|R^{sw}(\bm{\alpha}^{*}) - R^{gt} \|_1,
\end{equation} 
where $R^{gt}$ denotes the ground truth of the rotation. To sum up, IKOL is supervised by the weighted sum of the regression loss $L_{reg}$ and optimization loss $L_{opt}$, The total loss $L_{total}$ is derived as
\begin{equation}
L_{total} = L_{reg} + w_4 L_{opt}, 
\end{equation} 
where $w_4$ denotes the corresponding optimization loss weights.

The above optimization (Eq. \ref{loss1}) is highly nonconvex, which is hard to train together with the regression branch. As we analyzed before, existing optimization differentiation solvers, such as OptNet \cite{amos2017optnet} and CvxPylayer \cite{agrawal2019differentiable}, are not suitable the nonconvex optimization. In this paper, we develop a Gauss-Newton Differentiation to train the above nonconvex optimization with the regression branch under an end-to-end framework. 

\subsection{Gauss-Newton Differentiation}
As the forward model operator $f(\cdot)$ is nonlinear, optimization first requires linearizing this problem. Denotes $\bm{\alpha}_{k-1}$ as the swing angle at ${k-1}$th iteration. To this end, Taylor’s series expansion is used to approximate $f(\cdot)$ at $\bm{\alpha}_{k-1}$ in the Eq. \ref{forward}, i.e.:
\begin{equation}
f(\bm{\alpha}) \approx f(\bm{\alpha}_{k-1}) + J_{k-1} (\bm{\alpha} - \bm{\alpha}_{k-1}),
\end{equation} 
where $J_{k-1}$ denotes the Jacobian matrix calculated from the ($k-1$) th iteration. Here the Jacobian matrix $J_{k-1}$ is associated with the input of swing angle $\bm{\alpha}$, twist rotation $R^{tw}$, rest pose $T$ and regressed joints $P$. To ensure the whole process is differentiable, the key is to derive $J_{k-1}$ with analytical formulation in the kinematics tree. Detailed derivation for $J_{k-1}$ is provided in supplementary. 

Linearizing the Eq. \ref{forward} results in

\begin{equation}
\Delta \bm{\alpha}_k = \arg\min \limits_{\Delta\bm{\alpha}}  \frac{1}{2} \| J_{k-1} \Delta\bm{\alpha} -  \Delta P_{k-1}\| _2 ^2,
\label{linear}
\end{equation} 
where $\Delta P_{k-1} = P - f(\bm{\alpha} _{k-1})$ is the residue, and $\Delta \bm{\alpha}_k = \bm{\alpha} - \bm{\alpha} _{k-1}$ are the Gauss-Newton directions of $\bm{\alpha}_{k-1}$. Here Eq. \ref{linear} is a linear problem, and the Gauss-Newton directions are computed in each iteration:


\begin{equation}
\begin{aligned}
\Delta \bm{\alpha}_k &= - (J_{k-1}^{T}J_{k-1} + \sigma I)^{-1} J_{k-1}^{T} \Delta P_{k-1}\\
&= - H(\bm{\alpha}_{k-1},P) \Delta P_{k-1},
\end{aligned}
\label{Delta}
\end{equation} 
where $\sigma$ is a positive value to avoid ill-posed solutions and $H(\bm{\alpha}_{k},P) := (J_{k-1}^{T}J_{k-1} + \sigma I)^{-1} J_{k-1}^{T}$. Given an initial guess $\bm{\alpha}_0$, the swing angle is iteratively updated as follows: 

\begin{equation}
\bm{\alpha}_k = \bm{\alpha}_{k-1} + \Delta \bm{\alpha}_k.
\label{update}
\end{equation} 
Note that $\bm{\alpha}^{k-1}$ and $\Delta \bm{\alpha}^k$ are analytically differentiable. The whole process is summarized in Alg. \ref{GN-Diff_alg}. 



In each step, e.g. the $k$-th step, the swing angle is iteratively updated as Eq. \ref{update}, while as $ k \rightarrow \infty$, the local optimum of $\bm{\alpha}_*$ can be derived as follows:

\begin{equation}
\bm{\alpha}_* = \lim_{k \to\infty}\bm{\alpha}_{k} =  \lim_{k \to\infty} ( \bm{\alpha}_{k-1} + \eta_k \Delta \bm{\alpha}_k ).
\end{equation} 
In practice, the Gauss-Newton-based solver is a second-order method, which just needs several iterations to converge. We have conducted several experiment to illustrate the convergence speed in the following section. To better understand the behavior of the optimization loss $L_{opt}$, we examined its gradient with respect to the one of the input parameters $P$, that is, $\frac{ \partial  \bm{\alpha}_k }{\partial P } $


\begin{equation}
\frac{ \partial  \bm{\alpha} _{k} }{\partial P }= \frac{ \partial  \bm{\alpha} _{k-1} }{\partial P } + \frac{ \partial \Delta  \bm{\alpha} _{k} }{\partial P }.
\end{equation} 
Since $\Delta  \bm{\alpha} _{k} $ is derived from Eq. \ref{Delta}, its gradient can be calculated as:

\begin{equation}
\begin{aligned}
\frac{ \partial \Delta  \bm{\alpha} _{k} }{\partial P } &= - \frac{ \partial  H(\bm{\alpha}_{k-1}, P)  }{\partial P } \Delta P_{k-1} \\
&- H(\bm{\alpha}_{k-1}, P) \frac{ \partial  \Delta P_{k-1}  }{\partial P }.
\end{aligned}
\label{Delta_grad}
\end{equation} 


Note that the backward process in Alg. \ref{GN-Diff_alg} is only used for illustration which is implemented with PyTorch's automatic differentiation engine.

\begin{algorithm}[tb]
\begin{spacing}{1.1} 
\caption{The GN-Diff algorithm to differentiate IKOL}
\label{GN-Diff_alg}
\textbf{Input}: $\bm{\alpha}_0$\\
\textbf{Parameter}:$T$, $P$, $R^{tw}$, $f(\cdot)$, $\sigma$\\
\textbf{Output}: $\bm{\alpha}_*$ \\
\textbf{Forward}:
\begin{algorithmic}[1] 
\FOR{k = 1, 2, $\cdots$}
	\STATE  Compute $f(\bm{\alpha}_{k-1})$ and $J_{k-1}$ \\
	\STATE  Set $\Delta P_{k-1} = P - f(\bm{\alpha} _{k-1})$ \\
	\STATE  Compute $\Delta \bm{\alpha}_k$ using Eq. \ref{Delta} \\
	\STATE  Update $\bm{\alpha}_k = \bm{\alpha}_{k-1} + \Delta \bm{\alpha}_k $\\
\ENDFOR \\
\end{algorithmic}
\textbf{Backward}:
\begin{algorithmic}[1] 
\STATE $\frac{ \partial  \bm{\alpha} ^{0} }{\partial P } = - \frac{ \partial  H(\bm{\alpha}^{0}, P)  }{\partial P } \Delta P^{0} - H(\bm{\alpha}^{0}, P) \frac{ \partial  \Delta P^{0}  }{\partial P } $ \\
\FOR{k = 1, 2, $\cdots$}
   \STATE  Compute $\frac{ \partial \Delta  \bm{\alpha} _{k} }{\partial P }$ using Eq. \ref{Delta_grad} \\
   \STATE  Update $ \frac{ \partial  \bm{\alpha} _{k} }{\partial P }= \frac{ \partial  \bm{\alpha} _{k-1} }{\partial P } + \frac{ \partial \Delta  \bm{\alpha} _{k} }{\partial P }$\\
\ENDFOR
\end{algorithmic}
\textbf{Returned values}: 	$\bm{\alpha}_*, \frac{ \partial  \bm{\alpha} ^{*} }{\partial P }$
\end{spacing}
\end{algorithm}


\noindent\textbf{Comparison to Implicit Differentiation Methods.}
Existing differentiation solvers are usually based on the implicit function theorem, which, however, is a quite time-consuming process and cannot handle the nonconvex optimization problem. The 3D-HPS problem in this paper involves highly nonconvex mapping from the 3D skeleton and body shapes to the relative body-part rotations. Our GN-Diff iteratively linearizes the objective function to obtain closed form Gauss-Newton directions $\Delta \bm{\alpha}_{k}$ and the automatic differentiation can be directly applied for end-to-end training. 

\noindent\textbf{Comparison to SPIN.} Similar to our method, both the regression- and optimization-based modules are used in SPIN, where the optimization branch directly generates the supervision signal to train the regression one. 
Instead, IKOL treats the optimization problem  Eq. 5 as an optimization layer that is differentiated by GN-Diff for end-to-end training. Benefiting from the fast convergence and close-form solution at each iteration of the Gauss-Newton method, GN-Diff is quite fast and effective compare to the optimization used in SPIN. Moreover, GN-Diff is not very sensitive to the initialization, which makes our training process more stable than SPIN.  

\noindent\textbf{Comparison to Hybrik.}
HybrIK also considered to reduce the error between the regressed joints $P$ and the reconstructed joints $Q$. To achieve this, HybrIK relies on adaptively updating the newly reconstructed parent joints. However, HybrIK only reduces the error of children's joints in the kinematic tree if the parent joint is out of position, thus limiting the performance improvement. By contrast, our IKOL can significantly improve the 3D-HPS performance by iteratively optimizing all joints in the kinematic tree under an end-to-end differentiable framework.
\begin{table*}[t]
\centering
\renewcommand\arraystretch{1.1}
\begin{tabular}{l|l|lll|lll} 
\toprule[2pt]
\multirow{2}{*}{\makecell[c]{Method}} & \multirow{2}{*}{\makecell[c]{Backbone}} &\multicolumn{3}{l|}{\makecell[c]{3DPW}}   & \multicolumn{3}{l}{\makecell[c]{MPI-INF-3DHP}} \\
\cline{3-5} \cline{6-8}
                         &   & \makecell[c]{PVE $\downarrow$} & \makecell[c]{MPJPE $\downarrow$} &  \makecell[c]{PA-MPJPE $\downarrow$}  & \makecell[c]{PCK $\uparrow$} & \makecell[c]{AUC $\uparrow$} & \makecell[c]{MPJPE $\downarrow$}            \\ 
 \cline{1-8}
HMR \cite{kanazawa2018end}          & \makecell[c]{ResNet-50}       & \makecell[c]{-}      & \makecell[c]{130.0}  & \makecell[c]{81.3}     & \makecell[c]{72.9} & \makecell[c]{36.5} & \makecell[c]{124.2}             \\
Kolotouros et al. 2019  & \makecell[c]{ResNet-50} & \makecell[c]{- }     & \makecell[c]{-}      & \makecell[c]{70.2}     & \makecell[c]{-}   & \makecell[c]{-}    & \makecell[c]{-}                 \\
Arnab et al. 2019   & \makecell[c]{-}      & \makecell[c]{-}   & \makecell[c]{-}      &\makecell[c]{72.2}     & \makecell[c]{-}    & \makecell[c]{-}    & \makecell[c]{-}                 \\
SPIN     \cite{kolotouros2019learning}     & \makecell[c]{ResNet-50}    & \makecell[c]{116.4}  & \makecell[c]{96.9}   & \makecell[c]{59.2} & \makecell[c]{76.4} & \makecell[c]{37.1} & \makecell[c]{105.2}              \\
I2L-MeshNet  \cite{moon2020i2l}  & \makecell[c]{ResNet-50}    & \makecell[c]{110.1} & \makecell[c]{93.2}   & \makecell[c]{58.6}     & \makecell[c]{-}    & \makecell[c]{-}    & \makecell[c]{-}                  \\ 
HybrIK  \cite{li2021hybrik}   & \makecell[c]{ResNet-34}   & \makecell[c]{94.5}      & \makecell[c]{80.0}   & \makecell[c]{48.8}  & \makecell[c]{86.2} & \makecell[c]{42.2} & \makecell[c]{91.0}               \\
SPEC \cite{kocabas2021spec}   & \makecell[c]{ResNet-50}    & \makecell[c]{118.5}      & \makecell[c]{96.5}   & \makecell[c]{53.2}  & \makecell[c]{-} & \makecell[c]{-} & \makecell[c]{-}               \\
HMR-EFT (Joo et al. 2021)  & \makecell[c]{ResNet-50}   & \makecell[c]{98.7}      & \makecell[c]{85.1}   & \makecell[c]{52.2}  & \makecell[c]{-} & \makecell[c]{-} & \makecell[c]{-}               \\
PARE   \cite{kocabas2021pare}  & \makecell[c]{HRNet-w32}   & \makecell[c]{94.2}      & \makecell[c]{79.1}   & \makecell[c]{46.4}  & \makecell[c]{-} & \makecell[c]{-} & \makecell[c]{-}               \\
ROMP  \cite{sun2021monocular}  & \makecell[c]{HRNet-w32}    & \makecell[c]{93.4}      & \makecell[c]{76.7}   & \makecell[c]{47.3}  & \makecell[c]{-} & \makecell[c]{-} & \makecell[c]{95.11}               \\
BEV    \cite{sun2022putting} & \makecell[c]{HRNet-w32}   & \makecell[c]{92.3}      & \makecell[c]{78.5}   & \makecell[c]{46.9}  & \makecell[c]{-} & \makecell[c]{-} & \makecell[c]{-}               \\
METRO (Lin et al. 2021a)   & \makecell[c]{HRNet-w64}   & \makecell[c]{88.2}      & \makecell[c]{77.1}   & \makecell[c]{47.9}  & \makecell[c]{-} & \makecell[c]{-} & \makecell[c]{-}               \\
Mesh Graphormer (Lin et al. 2021b)  & \makecell[c]{-}     & \makecell[c]{87.7}      & \makecell[c]{74.7}   & \makecell[c]{45.6}  & \makecell[c]{-} & \makecell[c]{-} & \makecell[c]{-}               \\
\cline{1-8}
Ours(IKOL)   & \makecell[c]{ResNet-34} & \makecell[c]{86.4}      & \makecell[c]{73.3}   & \makecell[c]{45.5}  & \textbf{\makecell[c]{87.9}} & \textbf{\makecell[c]{48.1}} & \textbf{\makecell[c]{88.8}}             \\
Ours(IKOL*)   & \makecell[c]{ResNet-34} & \textbf{\makecell[c]{84.1}}      & \textbf{\makecell[c]{71.1}}   & \textbf{\makecell[c]{44.5}}  & \makecell[c]{87.0} & \makecell[c]{47.6} & \makecell[c]{89.7}               \\
\bottomrule[2pt]
\end{tabular}
\caption{Performance comparison on 3DPW and MPI-INF-3DHP. "*" means the results are fine tuned with the HybrIK model. "-" means the results that are not available. Best in \bf{bold}.}
\label{tab:benchmark}
\end{table*}

\section{Experiment}

\subsection{Implementation Details}
For the purposes of experimentation, we implemented our framework as follows. Although the optimization is perform iteratively through the network, we observed that, with the Gauss-Newton solver, only several steps were needed for the model to converge. Hence, we empirically  set the number of iteration steps to 5. It is worth noting that the objective function of our optimization only imposes a data fitting term; there is no regularization term. Therefore, unlike traditional optimization-based methods, our method is insensitive to the initialization parameter setting. We set the initialization parameter to zeros. Additionally, following \cite{li2021hybrik,moon2020i2l}, we used a ground truth bounding box in both the training and the testing stages. Mask R-CNN \cite{he2017mask} was used to derive the bounding box if one was not available in the dataset. Following HybrIK as our baseline, we used ResNet-34 \cite{he2016deep} pre-trained on the ImageNet dataset as our backbone network and adopted the same regression structure. All input images were padded to the same size of 256 × 256. We used Adam as our optimizer with an initial learning rate of $10^{-3}$, reduced by a factor of 10 at the 50th and the 150th epoch. The model was trained for 400 epochs with a mini-batch size of 64 per GPU on four RTX 3090 GPUs. Notably, we empirically found HybrIK is a great method to pretrain our model and our performance can further boost if we use this method as our initial model weight for training. Accordingly, we conduct two experiment with and without pretrain from HybrIK.

\begin{figure*}[t]
\centering
\includegraphics[scale = 0.6]{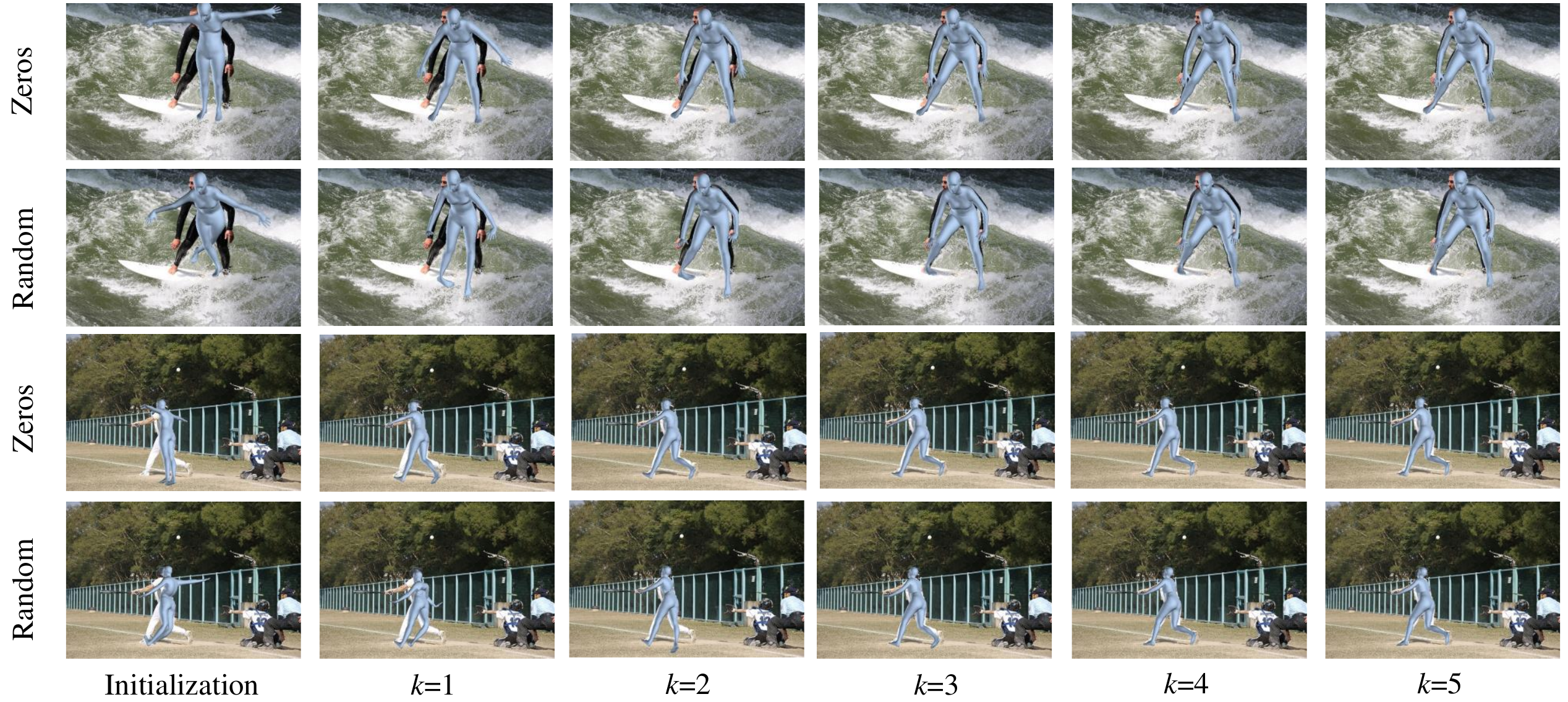}
\caption{Visualization of reconstruction output results across each Gauss-Newton iteration $k$ in the IKOL which show our method only take a few iterations to converge. Zero and random means use zero initialization and random initialization for fitting, which are used to verify the impact of the initialization}
\label{convergence}
\end{figure*}

\subsection{Datasets and Evaluation Metrics}
For a fair comparison with previous works, we trained our model on a mixture of 3D datasets (Human3.6M and MPI-INF-3DHP), and the 2D dataset COCO \cite{lin2014microsoft}. We used the 3DPW training data when conducting experiments on this dataset. For evaluation Metrics, mean per joint position error(MPJPE) and procrustes aligned mean per joint position error (PA-MPJPE) are used to reflect the performance of 3D pose. Besides, per Vertex Error (PVE) are to evaluate the body mesh estimation ability. Additionly, we report the Percentage of Correct Keypoints (PCK) and Area Under Curve (AUC) on the MPI-INF-3DHP dataset. Our main result is shown in Table \ref{tab:benchmark}.

\noindent\textbf{3DPW} 
3DPW is a challenging 3D-HPS benchmark that contains 60 video captured in indoor and outdoor conditions. We used this dataset as our major test set for evaluation. 

\noindent\textbf{MPI-INF-3DHP}
Following the setting of previous works, we report the MPJPE, the Percentage of Correct Keypoints (PCK) thresholded at 150mm and the Area Under the Curve (AUC) on MPI-INF-3DHP \cite{mehta2017monocular}


\noindent\textbf{Human3.6M Dataset.} The Human3.6M dataset \cite{ionescu2013human3} is the largest publicly available dataset for human 3D pose and shape estimation. Following \cite{li2021hybrik,moon2020i2l}, we used \emph{Protocol} 2 and sampled every 5th frame in the videos for training.


\begin{table}
\centering
\renewcommand\arraystretch{1.1}
\begin{tabular}{l<{\centering}|l<{\centering}|l<{\centering}|l<{\centering}} 
\toprule[2pt]
                        & \makecell[c]{PVE $\downarrow$} &\makecell[c]{ MPJPE $\downarrow$} & \makecell[c]{PA-MPJPE $\downarrow$}             \\ 
\cline{1-4}
Analytical solution       & \makecell[c]{94.5} & \makecell[c]{80.0}   & \makecell[c]{48.8}    \\ 
Our     & \textbf{\makecell[c]{92.5}}      & \textbf{\makecell[c]{78.2}}   & \textbf{\makecell[c]{48.5}}   \\
\bottomrule[2pt]
\end{tabular}
\caption{Ablation study of the iterative optimization in our IKOL compared to the analytical method in HybrIK on 3DPW dataset. Best in \bf{bold}.}
\label{tab:ablation-IKOL}
\end{table}


\subsection{Comparison with the State of the Art on Benchmark}
To compare our method with state-of-the-art, we report our quantitative results on 3DPW and MPI-INF-3DHP dataset. For fair comparison, we use 14 joints for 3DPW and 17 joints for MPI-INF-3DHP dataset following the previous method. Tabel \ref{tab:benchmark} shows that IKOL outperforms the existing regression and optimization methods in all sequence by a large margin. These results further demonstrate the effectiveness and efficiency of proposed IKOL model for 3D-HPS estimation.


\subsection{Ablation Study}
\noindent\textbf{Analytical method (HybrIK) VS Our iterative optimization}
To demonstrate the effectiveness of IKOL, we first compare the iterative optimization in our IKOL to the analytical method in HybrIK and take 3DPW dataset for an example to show the comparison. Since IKOL is completely compatible with the HybrIK module, we directly use the code from HybrIK and replace the analytical IK module in HybrIK with the iterative optimization in IKOL. It's worth noting that after the replacement we do not implement any fine tuning procedures to ensure a fair comparison. Tab. \ref{tab:ablation-IKOL} shows that IKOL achieve better performance by using the same checkpoints. This means that our method can be directly applied to HybrIK to further improve the performance.

\begin{table}
\centering
\renewcommand\arraystretch{1.1}
\begin{tabular}{l<{\centering}|l<{\centering}|l<{\centering}|l<{\centering}} 
\toprule[2pt]
                        & \makecell[c]{PVE $\downarrow$} & \makecell[c]{MPJPE $\downarrow$} & \makecell[c]{PA-MPJPE $\downarrow$}             \\ 
\cline{1-4}
Without GN-Diff       & \makecell[c]{91.5} & \makecell[c]{78.4}   & \makecell[c]{ 48.7 }    \\ 
Full model     & \textbf{\makecell[c]{86.4}}      & \textbf{\makecell[c]{73.3}}   & \textbf{\makecell[c]{45.5}}   \\
\bottomrule[2pt]
\end{tabular}
\caption{Ablation study of the GN-Diff design on the 3DPW dataset by ablating the optimization loss during training. Best in \bf{bold}.}
\label{tab:ablation-GNDiff}
\end{table}

\noindent\textbf{Analysis of the effectiveness of GN-Diff.}
To further verify the effectiveness of GN-Diff to train our network with optimization. We conduct the experiment on 3DPW dataset with and without GN-Diff during training. Two experimental setting were started from scratch. Table \ref{tab:ablation-GNDiff} shows that GN-Diff is able to improve the pose and shape estimation by 5.1 MPJPE, 3.2 PA-MPJPE and 4.6 PVE improvement.


\noindent\textbf{Analysis of initialization}
To show the robustness of our method, we conduct show the visualization of reconstruction output results across each iteration $k$ in the IKOL with different initialization as shown in Fig. \ref{convergence}. We use zero initialization and random initialization to verify that our method is insensitive to the initialization. From our empiricism, the whole training phase of IKOL is stable and we do not need any hand-craft thresholding to remove the bad fits as introduced in \cite{kolotouros2019learning}. Besides, although IKOL is insensitive to the initialization, we also need to keep our initialization parameter within reasonable values due to the non-convex of the object function. In practice, the absolute value of the initialization parameter is recommended to be set below 1. In our experiment, we set zero initialization parameters as default. 


\noindent\textbf{Time Cost of the Optimization Branch}
To analyze the time cost of the optimization branch in IKOL, we show some visualization results at each iteration $k$ in the IKOL. As shown in Fig. \ref{convergence}, we can see that IKOL only takes a few iterations to converge with different initialization. Further, we also compare the running time of our iterative optimization solver with the traditional optimization-based method SMPLify \cite{bogo2016keep} that is used in SPIN and analytical solution on a desktop computer with RTX 3080 GPU and Intel i7-10700 CPU. Table \ref{tab:runtime} shows that IKOL achieves superior performance, significantly faster than the competing optimization-based methods used in SPIN and even competitive with the analytical solution used in HybrIK without sacrificing too much computational cost. Note that other benchmark methods in Table \ref{tab:benchmark} do not involve the optimization branch, thus we do not need to compare with them here. 


\begin{table}
\centering
\renewcommand\arraystretch{1.1}
\begin{tabular}{l<{\centering}|l<{\centering}|l<{\centering}|l<{\centering}} 
\toprule[2pt]

                        & IKOL &  SMPLify  &     HybrIK  \\ 
                        \midrule
Runtime (s)       & \makecell[c]{0.182}  & \makecell[c]{74}   & \makecell[c]{0.04}   \\
\bottomrule[2pt]
\end{tabular}
\caption{Runtime comparisons on a 3080 GPU compared with the analytical solver of HybrIK and optimization scheme, SMPLify, used in SPIN. We set 5 iterations in our method for comparison.}
\label{tab:runtime}
\end{table}



\section{Conclusion}
In this paper, we presented IKOL, which leverages the strength of both optimization-based and regression-based methods within an end-to-end framework for 3D-HPS estimation. Unlike existing hybrid optimization and regression methods, IKOL involves an iterative optimization process within a DNN coupled with a regression branch. To ensure the framework offers end-to-end training, we designed GN-Diff to iteratively linearize the nonconvex objective function. As a result, the Gauss-Newton directions are derived in their closed form solutions, which means the differentiation procedure is both automatic and can be applied directly. In contrast to previous analytical derivations of inverse kinematics problems, IKOL boosts regression performance. And, in contrast to previous optimization and regression methods, our method simultaneously boosts the performance of both these popular paradigms.



\section{Acknowledgements}
This work was supported by the Shanghai Sailing Program (21YF1429400, 22YF1428800), Shanghai Local college capacity building program (22010502800), NSFC programs (61976138, 61977047), the National Key Research and Development Program (2018YFB2100500), STCSM (2015F0203-000-06), SHMEC (2019-01-07-00-01-E00003), and Shanghai Frontiers Science Center of Human-centered Artificial Intelligence (ShangHAI).

\bibliography{aaai23}

\end{document}